\documentclass[conference]{IEEEtran}
\IEEEoverridecommandlockouts
\usepackage{cite}
\usepackage{amsmath,amssymb,amsfonts}
\usepackage{algorithm}
\usepackage{algorithmic}
\usepackage{graphicx}
\usepackage{textcomp}
\usepackage{xcolor}
\usepackage{fancyhdr}
\usepackage{bm}
\usepackage{hyperref}
\usepackage{subfig}
\def\BibTeX{{\rm B\kern-.05em{\sc i\kern-.025em b}\kern-.08em
    T\kern-.1667em\lower.7ex\hbox{E}\kern-.125emX}}

\begin{document}

\title{Offloading and Quality Control for AI Generated Content Services in 6G Mobile Edge Computing Networks
}

\author{\IEEEauthorblockN{Yitong Wang, Chang Liu, Jun Zhao}
\IEEEauthorblockA{School of Computer Science and Engineering \\
Nanyang Technological University, Singapore\\
\{yitong002, liuc0063\}@e.ntu.edu.sg, junzhao@ntu.edu.sg}
}

\maketitle
\thispagestyle{fancy}
\pagestyle{fancy}
\lhead{This paper appears in the 2024 IEEE 99th Vehicular Technology Conference (VTC).\\ Please feel free to contact us for questions or remarks.}
\cfoot{\thepage}
\renewcommand{\headrulewidth}{0.4pt}
\renewcommand{\footrulewidth}{0pt}

\begin{abstract}
AI-Generated Content (AIGC), as a novel manner of providing Metaverse services in the forthcoming Internet paradigm, can resolve the obstacles of immersion requirements. Concurrently, edge computing, as an evolutionary paradigm of computing in communication systems, effectively augments real-time interactive services. In pursuit of enhancing the accessibility of AIGC services, the deployment of AIGC models (e.g., diffusion models) to edge servers and local devices has become a prevailing trend. Nevertheless, this approach faces constraints imposed by battery life and computational resources when tasks are offloaded to local devices, limiting the capacity to deliver high-quality content to users while adhering to stringent latency requirements. So there will be a tradeoff between the utility of AIGC models and offloading decisions in the edge computing paradigm. This paper proposes a joint optimization algorithm for offloading decisions, computation time, and diffusion steps of the diffusion models in the reverse diffusion stage. Moreover, we take the average error into consideration as the metric for evaluating the quality of the generated results. Experimental results conclusively demonstrate that the proposed algorithm achieves superior joint optimization performance compared to the baselines. 
\end{abstract}


\section{Introduction}

With GPT models capturing the spotlight, Generative Artificial Intelligence (GAI), as a transformative field within the broader landscape of machine learning and artificial intelligence, has changed the way people interact with and understand the digital world \cite{floridi2020gpt}. The demonstration of the capabilities inherent in Generative Artificial Intelligence (GAI) models is referred to as AI-generated content (AIGC). With the development of AIGC techniques, multiple AIGC models (e.g., diffusion models) can be employed to generate outputs with diverse forms, including text-to-speech, text-to-image, and image-to-image \cite{xu2023versatile}. Therefore, AIGC-as-a-Service (AaaS) architecture is proposed to offer the generated content, repair corrupted images, or alter inputted images, resulting in providing Metaverse users with immersive AIGC services. 

Edge computing, as a novel computing paradigm, not only solves the latency concerns of cloud computing but also improves the security and quality of the communication networks involved by integrating with other key technologies including AI, Blockchain, and digital twin\cite{mao2017survey}. The convergence of the AIGC models and edge computing paradigm becomes the focus of the future research direction, especially in the Metaverse field\cite{xu2023sparks} which emphasizes the immersion of users. 

6G, as the successor of the current 5G wireless communication network, aims to make further improvements in reliability, speed, and security, far surpassing the capabilities of 5G. The integration of 6G with edge computing is expected to revolutionize network architectures and computing, offering seamless, ultra-reliable, low-latency communication coupled with powerful, localized computing capabilities. This convergence will enable new applications and services that require high data rates, massive connectivity, and ultra-reliable low-latency communications, resulting in enhanced mobile broadband (eMBB), ultra-reliable low-latency communications (URLLC), and massive machine type communications (mMTC). 

Nonetheless, generic diffusion models (e.g., Open AI’s DALL-E 2 \cite{xie2023smartbrush} and Google’s Imagen \cite{saharia2022photorealistic}) require substantial memory storage, so deploying large AIGC models is challenging due to the large volume of parameters \cite{xu2023unleashing}, which brings obstacles for real-time applications and devices with constrained computational resources. Though lightweight diffusion models (like the text-to-image model with 860M UNet and 123M text encoder proposed in \cite{rombach2022high}) are employed on consumer mobile devices, the quality of generated content will also inevitably be affected. Thus, the convergence of diffusion models and edge computing systems has become the important direction of the future research area.

\textbf{Challenges:} In light of the resource constraints inherent in mobile devices, only lightweight diffusion models can be deployed on the local devices, resulting in the locally generated content typically exhibiting a lower quality level. While larger models can be deployed on the edge server, it remains incapable of simultaneously managing all computational tasks while generating high-quality content. Consequently, the first challenge centers on the offloading decisions and quality level of the generated content. Secondly, different locations to process the computational task means the computation time is always different. As mobile users possess diverse and stringent latency requirements, how to balance the tradeoff between the offloading decisions and computation time is the second challenge. Furthermore, the subjective nature of evaluating the quality level of generated content necessitates the establishment of a mathematical correlation between quality levels and diffusion models, thus constituting the third challenge.





\textbf{Related Work:} Recently, multiple studies have reviewed the state-of-the-art research and development in diffusion models \cite{croitoru2023diffusion,yang2022diffusion,zhang2023adding,li2023q,franzese2023much,zhang2023perceptual}. Notably, within the research of diffusion steps and image quality, \cite{franzese2023much} focuses on the forward diffusion dynamics to bridge the gap between the ideal and the simulated by adopting smaller diffusion times. Furthermore, \cite{chung2022come} treats the number of reverse diffusion steps as a variable to find the optimal reverse diffusion step to balance the tradeoff between the image quality and the diffusion time. For the convergence of the AIGC models and edge computing paradigm, \cite{10158526} introduces the diffusion model-based AI-generated optimal decision (AGOD) algorithm to provide the optimal strategy for the selection of AIGC service providers (AGPs). Unlike the previous works, we propose a resource allocation scheme for the AGOD by correlating the image quality to the diffusion steps and taking the computation time and utility of AIGC models into consideration. 

\textbf{Contributions:} The main contributions in this work are:\par
\begin{itemize}
    \item We are pioneering the implementation of the resource allocation for diffusion models in edge computing systems while guaranteeing the quality of the AIGC, which could enhance the performance and experience of the system and users respectively.
    \item We quantify the quality level of AICG by taking the reverse diffusion steps within the diffusion process into consideration while considering the average error of computation results. 
    \item The proposed optimization algorithm jointly optimizes the tradeoff between the offloading decisions of the computational tasks and the utility of AIGC models.
\end{itemize}

\section{System Model}

We consider a mobile edge computing communication system with $N$ mobile user equipments (UEs) and one edge server. Assume in this system that each UE requests to access AIGC services and content of corresponding quality level is generated at the chosen point based on the offloading decision and allocated reverse diffusion steps. 

\textbf{AI-generated content model.} To generalize the types of AIGC services (e.g. text-to-image, text-to-video, and image-to-image generation), we employ score-based diffusion models within our proposed system as the exemplar of AIGC models. Assume that there has been a standard sampling and diffusion process $S$ in the forward stage. AIGC services are then mainly provided by gradually reversing the diffusion process, step by step.  Consequently, the quality of AIGC computation results is intricately related to the high quantity of reverse diffusion steps during the reverse diffusion stage. 

When the generation requests of AIGC services are evaluated by the monitor, 
different offloading decisions and diffusion strategies of computational tasks are allocated to the edge server or local UEs based on the current energy conditions of the system. We denote the allocated offloading decisions and reverse diffusion steps of computation tasks requested by UE $n$ in the reverse stage as the binary variable $a_n\in \{0,1\}$ and discretization variable $s_n$ respectively. Specifically, $a_n=0$ indicates that UE $n$'s task will be processed locally on the mobile device, while $a_n=1$ indicates that the computation task will be offloaded to the edge server.

For the reverse diffusion step $s_n$, as elucidated in the work by Du \textit{et al.} \cite{10158526}, it exhibits a positive correlation with the associated energy expenditure. Given the finite energy resources inherent in the practical system, it is thus noteworthy that $s_n$ is restricted as each step of the reverse diffusion process necessitates energy consumption, primarily associated with the execution of a neural network for Gaussian noise removal. Therefore, we transform the total energy limitation into the constraint of total supported reverse diffusion steps at different servers. To enhance the generality of the proposed algorithm, we denote the total supported reverse diffusion steps at the edge server as $S_{e}^{\text{max}}$ with the constraint illustrated as:
\begin{align}
    \sum_{n\in \mathcal{N}}(a_ns_n)\leq S_{e}^{\text{max}},\label{p1c2}
\end{align}

where $\mathcal{N}=\{1,2,...,N\}$ is the set of UEs. Furthermore, we denote $S_{n,0}^{\text{max}}$ as the total supported reversed diffusion steps at each local UE $n$. In order to mitigate the potential for specific computation tasks to greedily consume computational resources on the edge server, a constraint is imposed wherein the maximum number of reverse diffusion steps allocated to each computation task is restricted. This constraint is denoted as the maximum reverse diffusion steps limit for each UE denoted by $S_{n,1}^{\text{max}}$, where $n$ signifies the $n$-th UE with $a_n=1$. Therefore, another constraint considering the limitation of energy including UE $n$ and the edge server is introduced as $(1-a_n)(s_n-S_{n,0}^{\text{max}})+a_n(s_n-S_{n,1}^{\text{max}}) \leq 0$, and after mathematical transformations, this constraint is expressed as:
\begin{align}
    s_n-(1-a_n)S_{n,0}^{\text{max}}-a_nS_{n,1}^{\text{max}}\leq 0, \forall n\in \mathcal{N}.\label{p1c3}
\end{align}


\subsection{Cost Functions}
\textbf{Computation time:} As per the findings reported in Page 6 of the paper \cite{chung2022come}, it is elucidated that the overall computation time can be expressed as an affine function with respect to the reverse diffusion steps. Consequently, the cumulative computational delays for tasks with different offloading decisions are determined as:
\begin{align}
    T_{n,0}(s_n)=s_n\Delta t_{n,0};~
    T_{n,1}(s_n)=s_n\Delta t_{n,1},
\end{align}
where $\Delta t_{n,0}$ and $\Delta t_{n,1}$ are designated as a constant temporal interval per individual step within the process at UE $n$ and the edge server. Given that the system is modeled for edge computing and the durations associated with transmitting the computed results through the downlink channels are negligible compared with the processing time of the edge server, it is pertinent to note that we shall exclude further consideration of downlink transmission time. 

\textbf{Average error of computation results:} 
In order to mitigate the influence of subjective factors on image quality and increase the alignment of the AIGC content with the users' request, we assess content quality by means of average error metrics. Building upon the Eq.(16) provided in the work \cite{chung2022come}, we derive that the reverse conditional diffusion pathway exhibits exponential error reduction and proposed a modified version which could be defined as:
\begin{align}
    \Bar{\epsilon}_n(s_n)=\Bar{\epsilon}^{\text{fwd}}(S)e^{-s_nC_{1,n}},
\end{align}
where function $\Bar{\epsilon}^{\text{fwd}}(S)$ is to determine the quality of processed content by adding Gaussian noise within the context of forward diffusion process, 
which can be modeled mathematically by a convex function related to the forward process $S$, and $C_{1,n}>0$, as the attenuation factor, represents the recovering ability of the AIGC model. When $S\rightarrow \infty$, it signifies that the 
forward diffusion has been sufficiently close to the unknown and simple noise distribution.  Furthermore, when $S\rightarrow 0$, there is no diffusion forward process and the average error converges to 0. Hence, function $\Bar{\epsilon}^{\text{fwd}}(S)$ increases as the forward diffusion process $S$ increases. 

\textbf{Total energy consumption:} Based on the relationship between CPU frequency (cycles/s) and data size (bits), and drawing from relevant work \cite{xiao2023joint}, the energy consumption associated with different offloading modes can be formulated as: When $a_n=0$, the energy consumption for UE $n$, denoted as $E_{n,0}(s_n)$, can be expressed as:
\begin{align}
    E_{n,0}(s_n)=k_nT_{n,0}(s_n)f_n^3,
\end{align}
where $f_n$ represents the CPU frequency of UE $n$, and $k_n$ is the coefficient reflecting the power efficiency of UE $n$. Conversely, when $a_n=1$, the energy consumption for UE $n$, referred to as $E_{n,1}(s_n)$, can be represented as 
\begin{align}
    E_{n,1}(s_n)=k_eT_{n,1}(s_n)g_n^3,
\end{align}
with $g_n$ representing the allocated computing capacity at the edge server for the computational task requested by UE $n$. Note that $k_e$ represents the analogous coefficient related to the power efficiency of the edge server.

\textbf{Cost functions:} Based on the previously delineated cost considerations, it can be deduced that, in the scenario where $a_n=0$, the cost function denoted as $\Bar{R}_{n,0}(s_n)$ is amenable to expression as $\Bar{R}_{n,0}(s_n)=c_1T_{n,0}(s_n)+c_2\Bar{\epsilon}_n(s_n)+c_3E_{n,0}(s_n)$. Conversely, in the case where $a_n=1$, the cost function assumes the form of $\Bar{R}_{n,1}(s_n)=c_1T_{n,1}(s_n)+c_2\Bar{\epsilon}_n(s_n)+c_3E_{n,1}(s_n)$. Herein, $c_1$, $c_2$, and $c_3$ denote the weighting coefficients for each cost component.

\subsection{Utility of Reverse Diffusion Steps}
We additionally contemplate the utility of reverse diffusion steps $s_n$ in the reverse diffusion process (i.e., the alignment of the generated content with the request). It is intuitive that the larger the reverse diffusion steps, the higher the diversity of content generated and the stronger the alignment with user requests. Consequently, the utility function should exhibit a non-decreasing relationship with respect to $s_n$. Besides, there is a marginal effect on the generation of the content (i.e., the content generation approaches saturation as $s_n$ increases), so the function should be concave. Then the utility function is defined as follows which is provided in \cite{8486241} for edge system:
\begin{align}
    U_n(s_n)=1-e^{-s_nC_{2,n}},
\end{align}
where $C_{2,n}\geq 0$ is the constant parameter. Variations in the parameter $s_n$ exert influence on the quality of the generated content, whereas diminished values of $s_n$ entail trade-offs in terms of energy consumption and utility. Therefore, it is imperative to establish a state of equilibrium.


\section{Joint Optimization of Cost and Utility}
In this section, we build the original optimization problem and make it convex by introducing auxiliary variables and adopting the penalized joint policy.

\subsection{Problem Formulation}

We define $\boldsymbol{a}:=[a_1,a_2,\ldots,a_N]$ and $\boldsymbol{s}:=[s_1,s_2,\ldots,s_N]$. In general, throughout this paper, for a vector $\boldsymbol{v}$, we denote its $i$-th dimension by $v_i$. A joint optimization problem, incorporating cost expenditure, offloading determinations, and utility is formulated as problem $\mathbb{P}_1$:\par
\vspace{-0.2cm}
\begin{small}
\begin{align}
    &(\mathbb{P}_1)\min_{\boldsymbol{a},\boldsymbol{s}}~\sum\nolimits_{n\in \mathcal{N}}\bigr[(1-a_n)R_{n,0}(s_n)+a_nR_{n,1}(s_n)\bigr] \label{p1} \\
    &\text{subject to}: a_n\in \{0,1\},\forall n \!\in \!\mathcal{N}, \label{p1c1}\\
    &~~~~~~~~~~~~~\text{Inequalities~(\ref{p1c2}) and~(\ref{p1c3}).} \nonumber
\end{align}
\end{small}%
where $R_{n, r}(s_n)=\omega_1\Bar{R}_{n, r}(s_n)-\omega_2U_n(s_n)$, $r\in \{0,1\}$, signifies the amalgamation of cost functions and utility functions, and $\omega_1$, $\omega_2$ serve as weight parameters specifically designated to modulate the magnitudes of the cost and utility components. $a_n$ and $s_n$ are optimization variables. Constraint (\ref{p1c1}) ensures the enforcement of binary offloading decisions. Constraint (\ref{p1c2}) imposes an upper bound on the total reverse diffusion steps $S_e^{\text{max}}$ for the edge server, thus addressing the constraint on available energy resources. Additionally, $S_{n,0}^{\text{max}}$ in constraint (\ref{p1c3}) serves as a mechanism for regulating the energy consumption of UE $n$, while $S_{n,1}^{\text{max}}$ acts as a constraint to prevent the excessive energy consumption associated with multiple computation tasks on the edge server. 

Then, we derived that the objective function of $\mathbb{P}_1$ is not jointly convex, as the binary value of the optimization variable 
$a_n$. Therefore, the optimization problem is stuck to be solved. More importantly, the discrete values of the variable $a_n$ and the coupling of the decision variables in the constraint (\ref{p1c2}) make this problem become an intractable NP-hard problem. 


\subsection{Transformation of the Non-Convex Problem}

Derived from the formulated objective function, it becomes evident that the terms associated with $a_n$ are amenable to extraction through standard factorization procedures. Henceforth, we employ a strategic approach involving the decoupling of the optimization variables, subsequently leading to the transformation of the initial problem $\mathbb{P}_1$ into a set of sub-problems \cite{sun2016majorization} that can be addressed iteratively.

\textbf{Handling objective function:} To tackle the coupling of optimization variables, surrogate functions of objective function can be constructed by introducing auxiliary variables \cite{10368052}. In the $(k+1)$-th iteration, the surrogate function is expressed as:
\begin{align}
    G(\bm{a},\bm{s}|\bm{u}^{(k)},\bm{v}^{(k)})=\sum\nolimits_{n\in \mathcal{N}}g_n(a_n,s_n|u_n^{(k)},v_n^{(k)}),
\end{align}
where $u_n^{(k)}=\frac{(1-a_n^{(k)})}{2R_{n,0}(s_n^{(k)})}$ and $v_n^{(k)}=\frac{a_n^{(k)}}{2R_{n,1}(s_n^{(k)})}\in \mathbb{R}^+$ are the introduced auxiliary variables related to the convergence of our algorithm which will be analyzed in Section \ref{3F}. Function $g_n(a_n,s_n|u_n^{(k)},v_n^{(k)})$ for UE $n$ is defined as:\par
\vspace{-0.3cm}
\begin{small}
\begin{align}
    g_n(\!a_n,\!s_n\!|u_n^{(k)}\!\!,\!v_n^{(k)})\!\!=\!\!R_{n,0}^2(s_n)u_n^{(k)}\!\!\!+\!\frac{(1-a_n)^2}{4u_n^{(k)}}\!\!+\!\!R_{n,1}^2(s_n)v_n^{(k)}\!\!\!+\!\!\frac{a_n^2}{4v_n^{(k)}}.\nonumber
\end{align}
\end{small}%

\textbf{Handling constraint (\ref{p1c2}):} In order to decouple the constraint (\ref{p1c2}), we adopt a similar approach by introducing an auxiliary variable $\bm{z}=[z_1,...,z_N]$ to facilitate the formulation of the surrogate function for each individual sub-problem. Consequently, in the $(k+1)$-th iteration, the constraint undergoes a transformation as follows: 
\begin{align}
    -S_{e}^{\text{max}}+\sum\nolimits_{n\in \mathcal{N}}(s_n^2z_n^{(k)}+\frac{a_n^2}{4z_n^{(k)}})\leq 0, 
\end{align}
where $z_n^{(k)}=\frac{a_n^{(k)}}{2s_n^{(k)}}$ constrains the variables in each loop and converges with iterations.

Noting that the coupling among the optimization variables has been addressed, it is pertinent to acknowledge that the optimization sub-problems remain non-convex due to the discrete nature of the variables $a_n$. 

\subsection{Successive Convex Approximation}
In this section, we focus on using the penalized joint policy and successive convex approximation (SCA) technique to handle the discrete variables $a_n$. 

\textbf{Handling constraint (\ref{p1c1}):} To solve the discrete variable $a_n$ and without loss of equivalence, it can be rewritten as:
\begin{align}
    &~~~~a_n\in [0,1], n\in \mathcal{N},\label{p1c1_1} \\ 
    &\sum\nolimits_{n \in \mathcal{N}}a_n(1-a_n) \leq 0. \label{p1c1_2}
\end{align}

\addtolength{\topmargin}{0.07in}

Note that the optimization problem has been transitioned into a continuous optimization problem, resulting in a notable reduction in computational complexity when contrasted with the direct resolution of the original discrete variable $a_n$. Nonetheless, the function $\sum_{n \in \mathcal{N}}a_n(1-a_n)$ in constraint (\ref{p1c1_2}) is a concave function. To further facilitate the solution, we adopt a method that introduces a penalty term for this concave constraint into the function $G(\bm{a},\bm{s}|\bm{u}^{(k)},\bm{v}^{(k)})$, represented as:\par
\vspace{-0.3cm}
\begin{small}
\begin{align}
    G(\bm{a},\bm{s}|\bm{u}^{(k)},\bm{v}^{(k)})-\tau\cdot \sum\nolimits_{n \in \mathcal{N}}a_n(a_n-1),
\end{align}
\end{small}%
where $\tau$ is the penalty parameter with $\tau >0$. Then the objective function becomes concave due to the concavity of the second term. Simultaneously, given the second term is differentiable, we utilize the first-order Taylor series to linearize it at each iteration. Specifically, at the $(i+1)$-th iteration,
we approximate $\sum_{n \in \mathcal{N}}a_n(a_n-1)$ with $\sum_{n \in \mathcal{N}}a_n^{(i)}(a_n^{(i)}-1)+(2a_n^{(i)}-1)(a_n-a_n^{(i)})$ which is denoted as $H(\bm{a}|\bm{a}^{(i)})$, where $\bm{a}^{(i)}$ is defined as the optimal solution of the $i$-th sub-problem. Consequently, the objective function is converted to:
\begin{align}
    G(\bm{a},\bm{s}|\bm{u}^{(k)},\bm{v}^{(k)})-\tau\cdot H(\bm{a}|\bm{a}^{(i)}).
\end{align}

Then, the $(i+1)$-th sub-problem in the $(k+1)$-th iteration is transformed equivalently to $\mathbb{P}_2$:
\begin{subequations} \label{p2}
\begin{align}
    &(\mathbb{P}_2)\min_{\boldsymbol{a},\boldsymbol{s}}~G(\bm{a},\bm{s}|\bm{u}^{(k)},\bm{v}^{(k)})-\tau\cdot H(\bm{a}|\bm{a}^{(i)}) \tag{\ref{p2}} \\[-2pt]
    &\text{subject to}:a_n\in [0,1], \forall n\in \mathcal{N}, \label{p2c1} \\[-2pt]
    &~~~~~~~~~~~~~\sum\nolimits_{n\in \mathcal{N}}(s_n^2z_n^{(k)}+\frac{a_n^2}{4z_n^{(k)}})-S_{e}^{\text{max}}\leq 0,\label{p2c2}\\[-2pt]
    &~~~~~~~~~~~~~s_n\!-\!(1-a_n)S_{n,0}^{\text{max}}-a_nS_{n,1}^{\text{max}}\leq 0, \forall n\in \mathcal{N}. \label{p2c3}
\end{align}
\end{subequations}

Until now, the original optimization problem $\mathbb{P}_1$ can be solved by using $\mathbb{P}_2$ iteratively, and the process of solving the intra-sub-problem is listed in Algorithm \ref{intra}. Given introduced conditions and the objective function are convex, $\mathbb{P}_2$ is convex. Thus, we can adopt  Karush-Kuhn-Tucker (KKT) conditions of $\mathbb{P}_2$ to obtain the optimal solutions. 

\begin{algorithm}[t]
    \renewcommand{\algorithmicrequire}{\textbf{Input:}}
    \renewcommand{\algorithmicensure}{\textbf{Output:}}
    \caption{Intra-Sub-Problem Programming}
    \label{intra}
    \begin{algorithmic}[1]
        \STATE Initialize: $k=0$; optimization variable $\bm{\mathcal{X}}^{(0)}\!=\![\bm{a}^{(0)}\!,\bm{s}^{(0)}]$; 
        \vspace{-0.4cm}
        \STATE Calculate and derive auxiliary variable space: $\bm{\mathcal{A}}^{(0)}=[\bm{u}^{(0)}\!\!,\bm{v}^{(0)}\!\!,\bm{z}^{(0)}]$, where $u_n^{(0)}\!=\frac{(1-a_n^{(0)})}{2R_{n,0}(s_n^{(0)})}$, $v_n^{(k)}\!=\frac{a_n^{(0)}}{2R_{n,1}(s_n^{(0)})}$, and $z_n^{(0)}=\frac{a_n^{(0)}}{2s_n^{(0)}}$, $\forall n\in \mathcal{N}$;
        \REPEAT
        \STATE Obtain the optimal variable $\bm{\mathcal{X}}^{(k+1)}$ of $((k+1))$-th iteration by adopting the Algorithm \ref{solveKKT} when given auxiliary variable space $\bm{\mathcal{A}}^{(k)}$;
        \STATE Update $\bm{\mathcal{A}}^{(k+1)}=[\bm{u}^{(k+1)}\!\!,\bm{v}^{(k+1)}\!\!,\bm{z}^{(k+1)}]$ with given $\bm{\mathcal{X}}^{(k+1)}$;
        \STATE $k\leftarrow k+1$;
        \UNTIL Convergence or reach the max iteration number $K$.
    \end{algorithmic}
\end{algorithm}

\subsection{KKT Conditions for Problem $\mathbb{P}_2$}

We first write down the Lagrange function of $\mathbb{P}_2$ by introducing multipliers for the constraints:
\begin{footnotesize}
\begin{align}
    &L\!=\!G(\bm{a},\bm{s}|\bm{u}^{(k)}\!\!,\bm{v}^{(k)})\!-\!\tau\!\!\cdot\!\! H(\bm{a}|\bm{a}^{(i)})\!+\!\!\sum_{n\in \mathcal{N}}\![\beta_n (\!-a_n)\!+\!\gamma_n(a_n\!\!-\!\!1)]\!+\!\delta\cdot\nonumber \\
    & \bigr[\!\!\sum_{n\in \mathcal{N}}\!\!(s_n^2z_n^{(k)}\!+\!\frac{a_n^2}{4z_n^{(k)}}\!)
    \!-\!S_{e}^{\text{max}}\bigr]\!\!+\!\!\!\sum_{n\in \mathcal{N}}\!\!\zeta_n\bigr[s_n\!\!-\!(\!1\!-\!a_n\!)S_{n,0}^{\text{max}}\!-\!a_nS_{n,1}^{\text{max}}\bigr], 
\end{align}
\end{footnotesize}

After applying KKT conditions, we get:

\textbf{Stationarity:}
\begin{footnotesize}
\begin{subequations}
\begin{align} 
    &\frac{\partial L}{\partial a_n}=D_n(a_n,\zeta_n,\delta)-\beta_n+\gamma_n =0, \\
    &\frac{\partial L}{\partial s_n}=\frac{\partial G(\bm{a},\bm{s}|\bm{u}^{(k)},\bm{v}^{(k)})}{\partial s_n}+2\delta s_n z_n^{(k)}+\zeta_n =0. 
\end{align} \label{kktS}
\end{subequations}
\end{footnotesize}
where $D_n(a_n,\zeta_n,\delta)=(\frac{1}{2u_n^{(k)}}+\frac{1}{2v_n^{(k)}}+\frac{\delta}{2z_n^{(k)}})a_n-\frac{1}{2u_n^{(k)}}-\tau (2a_n^{(i)}-1)+\zeta_n(S_{n,0}^{\text{max}}-S_{n,1}^{\text{max}})$ relating to the variable $a_n$. 

\textbf{Complementary slackness:}
\begin{footnotesize}
\begin{equation}
\begin{split}\label{kktC}
    &\text{(\ref{kktC}a)}\!:\!\beta_n(\!-a_n\!)\!=\!0; \hspace{7pt}\text{(\ref{kktC}c)}\!:\!\delta \bigr[\!\!\sum_{n\in \mathcal{N}}\!\!(s_n^2z_n^{(k)}\!+\!\frac{a_n^2}{4z_n^{(k)}}\!)
    \!-\!S_{e}^{\text{max}}\bigr]\!=\!0;  \\
    &\text{(\ref{kktC}b)}\!:\!\gamma_n(\!a_n\!-\!1\!)\!=\!0; \hspace{2pt}\text{(\ref{kktC}d)}\!:\!\zeta_n\bigr[s_n\!\!-\!(\!1\!-\!a_n\!)S_{n,0}^{\text{max}}\!-\!a_nS_{n,1}^{\text{max}}\bigr]\!=\!0. 
\end{split}
\end{equation}
\end{footnotesize}

\textbf{Primal Feasibility:} \text{(\ref{p2c1})}, (\ref{p2c2}), (\ref{p2c3}). \label{kktP}

\textbf{Dual Feasibility:} 
\begin{equation}
\begin{split} \label{kktD}
    \text{(\ref{kktD}a)-(\ref{kktD}d)}: \beta_n, \gamma_n, \delta, \zeta_n \geq 0, \forall n \in \mathcal{N}. 
\end{split}
\end{equation}

Under the aforementioned conditions, we then proceed to seek the optimal solutions by analyzing the KKT conditions and employing the proposed algorithm which is listed in Algorithm \ref{solveKKT}. 

\begin{algorithm}[t]
    \renewcommand{\algorithmicrequire}{\textbf{Input:}}
    \renewcommand{\algorithmicensure}{\textbf{Output:}}
    \caption{Solve KKT Conditions}
    \label{solveKKT}
    \begin{algorithmic}[1]
        \STATE Given the auxiliary variable space $\bm{\mathcal{A}}=[\bm{u},\bm{v},\bm{z}]$;
        \FOR{$n\leftarrow 1 \text{ to } N$}
            \STATE Obtain $\widehat{a}_n(\zeta_n,\delta)$ by assuming $D_n(a_n,\zeta_n,\delta)$ on condition (\ref{kktS}a) equal to 0;
            \STATE Obtain $\widetilde{s}_n(\zeta_n,\delta)$ based on condition (\ref{kktS}b);
            \STATE Obtain $\widehat{\zeta}_n(\delta)$ when assuming $I_n(\zeta_n,\delta)=0$ in the condition (\ref{kktC}d);
        \ENDFOR
        \STATE $\delta^*\!\!\leftarrow\! \!\left \{ \!\! \!
         \begin{small}
         \begin{array}{l}
            0, \text{if} \sum_{n\in \mathcal{N}}(\widetilde{s}_n^2(\widetilde{\zeta}_n(0),0)z_n^{(k)}+\frac{a_n^2(\widetilde{\zeta}_n(0),0)}{4z_n^{(k)}})\leq S_{e}^{\text{max}}; \\
            \text{Solution to } \sum_{n\in \mathcal{N}}(\widetilde{s}_n^2(\widetilde{\zeta}_n(\delta)\!,\!\delta)z_n^{(k)}\!\!\!+\!\frac{a_n^2(\widetilde{\zeta}_n(\delta),\delta)}{4z_n^{(k)}}\!)\!\!=\! \!S_{e}^{\text{max}};
         \end{array}
         \end{small}
         \right. $
        \STATE \textbf{update} $~a_n^*\!\leftarrow\!\max\{\min\{\widehat{a}_n(\widetilde{\zeta}_n(\delta^*),\delta^*),1\},0\}$, and $s_n^*\!\!\leftarrow\widetilde{s}_n(\widetilde{\zeta}_n(\delta^*),\delta^*)$;
        \RETURN The optimal variable value  $\mathcal{X}^*=[\bm{a}^*,\bm{s}^*]$.
    \end{algorithmic}
\end{algorithm}

\textbf{Theorem 1:} \textit{The optimal solution of the proposed objective function can be obtained by \textbf{Algorithm \ref{solveKKT}} and is expressed as:}
\begin{align}
    \left \{\!\!
    \begin{array}{l}
         a_n^*=\max\{\min\{\widehat{a}_n(\widetilde{\zeta}_n(\delta^*),\delta^*),1\},0\};\\
         s_n^*=\widetilde{s}_n(\widetilde{\zeta}_n(\delta^*),\delta^*).
    \end{array}
    \right.
\end{align}
where $\widetilde{\zeta}_n(\delta)=\max\{\widehat{\zeta}_n(\delta),0\}$. More specifically,  $\widehat{a}_n(\zeta_n,\delta)$ meets the condition (\ref{kktS}a) with $D_n(a_n,\zeta_n,\delta)|_{a_n=\widehat{a}_n(\zeta_n,\delta)}\!=\!0$ and $\widehat{\zeta}_n(\delta)$ satisfies condition (\ref{kktC}d) with $I_n(\zeta_n,\delta)|_{\zeta_n=\widehat{\zeta}_n(\delta)}\!=\!0$.

\textit{Proof:} Please see Appendix \ref{kktanalysis}. $\hfill\blacksquare$

\subsection{Inter-Sub-Problem Algorithm}
After completing the analysis of KKT conditions within each sub-problem, we obtain the optimal solution for each sub-problem. In this section, we introduce the inter-sub-problem algorithm, denoted as Algorithm \ref{inter}, which leverages iterative implementations of the SCA method. Its primary objective is the determination of the globally optimal solution for the original optimization problem $\mathbb{P}_1$.

\begin{algorithm}[t]
    \renewcommand{\algorithmicrequire}{\textbf{Input:}}
    \renewcommand{\algorithmicensure}{\textbf{Output:}}
    \caption{Inter-Sub-Problem Programming}
    \label{inter}
    \begin{algorithmic}[1]
        \STATE Initialize: $i=1$; Optimal solution space $\bm{\mathcal{S}}^{(0)}\!\!\!=\!\![\bm{a}^{(0)}\!\!,\bm{s}^{(0)}]$; 
        \STATE Adopt SCA method to obtain $\mathbb{P}_2$ from $\mathbb{P}_1$;
        \REPEAT
        \STATE Obtain the $i$-th sub-problem by using $\bm{a}^{(i-1)}$;
        \STATE Solve the $i$-th sub-problem by using Algorithm \ref{intra} and \ref{solveKKT} to get the optimal optimal variable value $\mathcal{X}^*
        \!=\![\bm{a}^*\!,\bm{s}^*]$;
        \vspace{-0.4cm}
        \STATE Let $\bm{\mathcal{S}}^{i}[\bm{a}^{(i)},\bm{s}^{(i)}]\leftarrow \mathcal{X}^*
        \!=\![\bm{a}^*\!,\bm{s}^*]$ of $i$-th sub-problem;
        \STATE Set $i\leftarrow i+1$;
        \UNTIL $|\bm{\mathcal{S}}^{(i)}-\bm{\mathcal{S}}^{(i-1)}|\leq \bar{\epsilon}_0$ or reach the maximum iteration number $I$;
        \RETURN $\bm{\mathcal{S}}^{(i)}[\bm{a}^{(i)},\bm{s}^{(i)}]$ as the optimal solution of $\mathbb{P}_1$.
    \end{algorithmic}
\end{algorithm}

\subsection{Time complexity, Solution Quality and Convergence} \label{3F}

\textbf{Time Complexity.} Based on the algorithms listed, the complexity of Algorithm \ref{inter} lies in the step 3-8. Assuming $K$ is the number of iterations in Algorithm \ref{inter}, then the complexity of adopting $\bm{a}^{i-1}$ to obtain the sub-problem in step 4 is denoted as $\mathcal{O}(IN)$, where $N$ derives from computing $a_n^{i-1}$ for each user in an iteration. In step 5, Algorithm \ref{intra} is mainly solved by Algorithm \ref{solveKKT} to obtain the optimization solution of the intra-sub-problem, so the complexity of Algorithm \ref{solveKKT} is first analyzed. The total complexity of Algorithm \ref{solveKKT} is $\mathcal{O}(2N)$ as the complexity of computing $\bm{a}^*$ and $\bm{s}^*$ are both $\mathcal{O}(N)$ separately. Let $K$ denote the number of iterations in Algorithm \ref{intra}. The complexity of the Algorithm \ref{intra} is thus $\mathcal{O}(3KN)$ as step 5 of Algorithm \ref{intra} also costs $\mathcal{O}(N)$. Therefore, the overall complexity can be derived as $\mathcal{O}((I+3KI)N)$.

\textbf{Solution quality and convergence.} Algorithm \ref{inter} comprises SCA, Algorithm \ref{intra}, and Algorithm \ref{solveKKT}. Though the SCA method adopted to transform $\mathbb{P}_1$ to $\mathbb{P}_2$ results in some loss of the optimality, the KKT analysis by introducing the auxiliary variables in Algorithm \ref{intra} and Algorithm \ref{inter} both are without loss of optimality. Thus Steps 3-8 of Algorithm \ref{inter} can guarantee to find the global optimal solutions for $\mathbb{P}_2$. The convergence of Algorithm 3 is also evident from the preceding analysis.

\section{Experimental Results}

In this section, we evaluate the prior performance of our proposed algorithm. Firstly, we introduce the numerical parameters settings and then discuss experimental results. 

\subsection{Parameter Settings}

In this experiment, we set the total number of users $N$ is 30. To generalize the experimental result without any distortion and error, the original value of the average error rate is set as 1, which implies the original content is complete Gaussian noise. Based on the resource limitation, the fixed discretization steps are $\Delta t_{n,0}=1/500$ and $\Delta t_{n,1}=1/1000$. The computing capacity $g_n$ and frequency of mobile user $f_n$ are set as $10$GHz and $1.5$GHz by default. The computation energy efficiency coefficient $k_n$ and $k_e$ is $10^{-26}$. For the penalty parameter $\tau$, we set the value as $10^5$.
\subsection{Performance when Adapting Weight Parameters}

\begin{figure*}[t]
    \centering
    \begin{minipage}{0.65\textwidth}
        \centering
        \subfloat{\includegraphics[width=0.32\textwidth,height=0.3\textwidth]{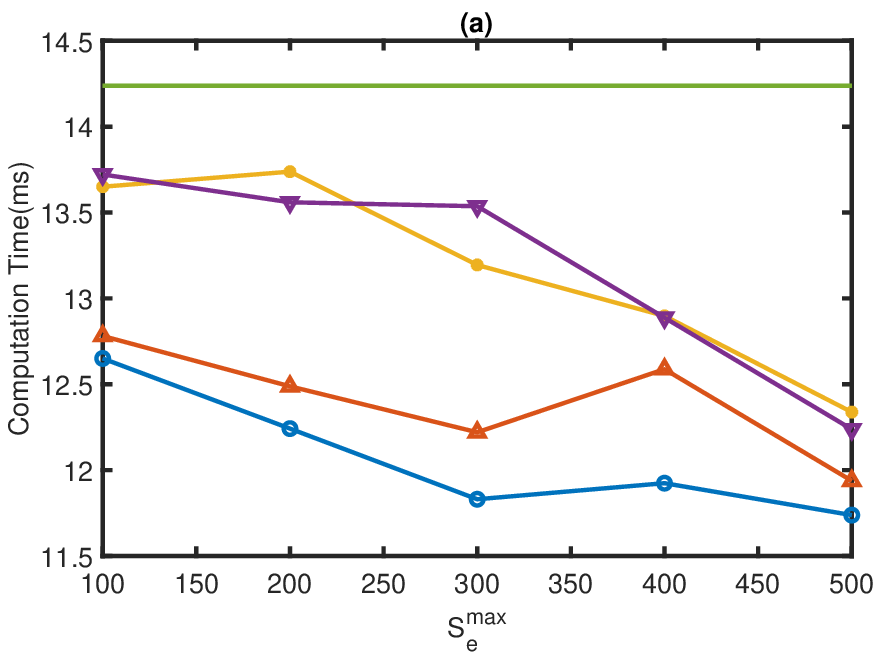}}
        \hspace{0.05in}
        \subfloat{\includegraphics[width=0.32\textwidth,height=0.3\textwidth]{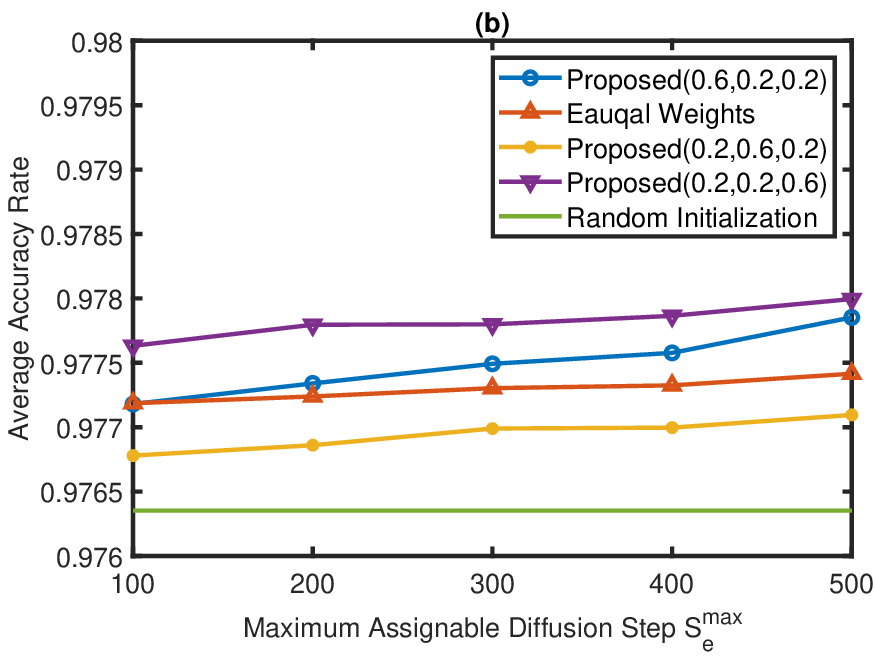}}
        \hspace{0.05in}
        \subfloat{\includegraphics[width=0.32\textwidth,height=0.3\textwidth]{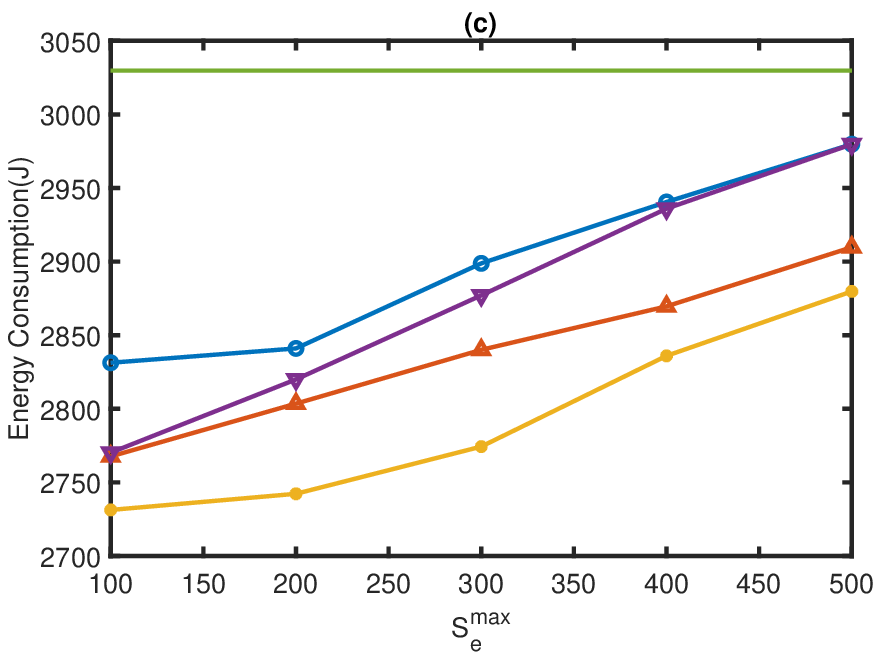}}
        \caption{Consumption under different maximum diffusion steps $S_e^{\max}$.}\vspace{-0.7cm}
        \label{step}
    \end{minipage}
    \hspace{0.1cm}
    \begin{minipage}{0.33\textwidth}
        \vspace{0.45cm}
        \includegraphics[width=1\textwidth,height=0.58\textwidth]{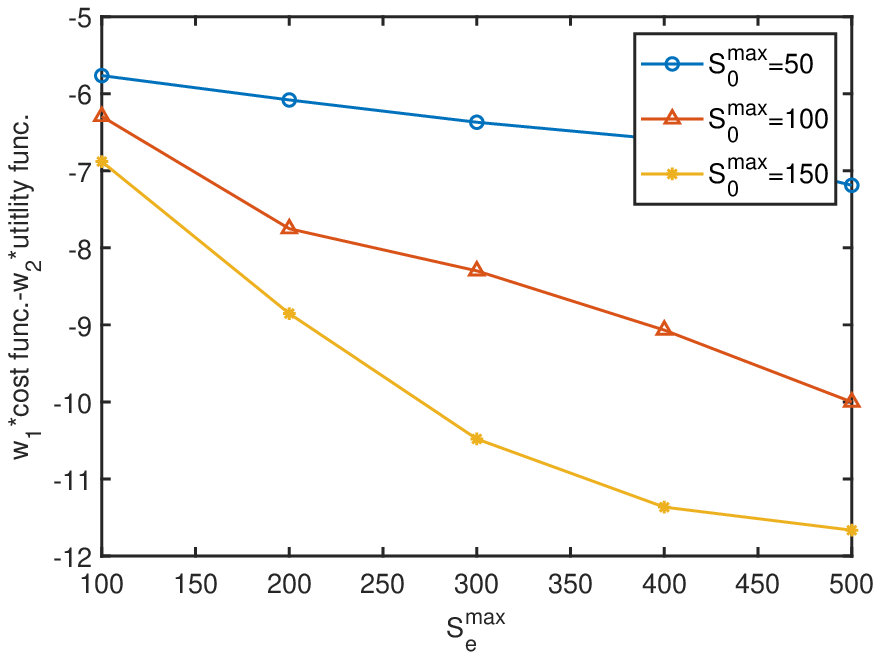}
        \vspace{-0.45cm}
        \caption{Joint optimization Performance.}
        \label{performence}\vspace{-0.7cm}
    \end{minipage}
\end{figure*}


As the weight factors $(c_1,c_2,c_3)$ of cost functions have an effect on the optimization process, we make modifications to these weights to tailor the focus of the associated cost functions. To investigate the different costs in the system, we augment $c_1$, $c_2$, and $c_3$ to respectively heighten the sensitivity of computational time, average error rate, and energy consumption. To further investigate the outcomes of our proposed algorithm, we conduct experimental trials across the combinations of $(c_1, c_2, c_3)$ and subsequently compare them to a baseline: random initialization. This random initialization entails the arbitrary allocation of diffusion steps following the stochastic selection of offloading decisions. For the baseline, we assume that no preference exists in the algorithm, then the weight factor is equal. Results of Fig.~\ref{step} show the computation $T$, average error of computation results $\Bar{\epsilon}$ and energy consumption $E$ under different maximum assignable diffusion step $S_e^{\max}$. From Fig.~\ref{step}(a),  we can see the total consumption time is decreased as $S_e^{\max}$ increase, which means more complex computational tasks are processed on the edge server and the fixed discretization step on the edge server is faster. As the energy consumption is also larger than the local devices, the energy consumption also increases with increasing reverse diffusion steps which is presented in Fig. \ref{step}(c). To provide a more intuitive representation of the quality of AIGC, we adopt the average accuracy ($1-\Bar{\epsilon}$) instead of average error as shown in Fig.~\ref{step}(b). For the different weights lines, the red line (proposed equal weights) outperforms others comprehensively as no sacrifice exists in this scenario compared with other lines. Furthermore, we can see that all proposed methods outperform the baseline, demonstrating our method's superiority.

\subsection{Performance when Setting Different Diffusion Steps}


To conduct a comprehensive investigation into the performance of the proposed method, we consider diverse joint optimization scenarios when the system requires different $S_e^{\max}$. To emphasize the utility function, we set the parameters $(w_1,w_2)$ as $(0.3,0.7)$. Three distinct experiments are undertaken, each characterized by different local reverse diffusion steps $S_0^{\max}$. As illustrated in Fig.\ref{performence}, we can see that as $S_e^{max}$ increases, all three lines exhibit an initial decline followed by a stabilization phase upon reaching an optimal solution. This observation underscores that as the system benefits from a higher resource allocation for task processing, it tends to prioritize the augmentation of the utility function to achieve superior joint optimization performance.

\section{Conclusion}

In this paper, we propose one joint optimization algorithm that bridges the gap between AIGC models and edge computing while mitigating the constraints posed by resource limitations on devices. The presented algorithm offers an enhanced approach by simultaneously optimizing offloading decisions and the reverse diffusion steps of diffusion models, taking into account average error and energy consumption. Our analysis of experimental results demonstrates the effectiveness of the proposed algorithm in enhancing the system efficiency.

\section*{Acknowledgement}

This research is supported partly by the Singapore Ministry of Education Academic Research Fund under Grant Tier 1 RT5/23, Grant Tier 1 RG90/22, Grant Tier 1 RG97/20, Grant Tier 1 RG24/20 and Grant Tier 2 MOE2019-T2-1-176; partly by the Nanyang Technological University (NTU)-Wallenberg AI, Autonomous Systems and Software Program (WASP) Joint Project; and partly by Imperial-Nanyang Technological University Collaboration Fund INCF-2024-008.

\bibliographystyle{IEEEtran}
\bibliography{reference.bib}

{
\appendix{\textbf{Analysis of KKT conditions:}} \label{kktanalysis}\
In order to identify the optimal $(\bm{a}^*, \bm{s}^*)$ that satisfies the KKT conditions, we present the ensuing analysis rooted in principles of optimization theory.  

Deriving from the functional expression $D_n(a_n,\zeta_n,\delta)$ as delineated in condition (\ref{kktS}a), two properties of condition (\ref{kktS}a) can be inferred contingent upon the values of auxiliary variables and Lagrange multipliers: \textit{1). $D_n(a_n,\zeta_n,\delta)$ is non-decreasing for $a_n$ and 2). Specifically, we obtain the explicit expression of $a_n$ by setting $\beta_n=0$ and $\gamma_n=0$.} Then we denote the explicit expression as $\widehat{a}_n(\zeta_n,\delta)$ and proceed to discuss the different cases based on the condition (\ref{kktC}a) and (\ref{kktC}b) as outlined below: 
\begin{itemize}
    \item \textbf{Case 1:} $\widehat{a}_n(\zeta_n,\delta)\geq  1$. In this case, we can infer that $D_n(1,\zeta_n,\delta) \leq 0$ based on the given two properties. Therefore, we can set $\widehat{a}_n(\zeta_n,\delta)=1$ and $\beta_n=0$ to meet the conditions, and then the value of $\gamma_n$ is equal to $-D_n(1,\zeta_n,\delta)\geq 0$ which exactly meet the condition (\ref{kktC}b). 
    \item \textbf{Case 2:} $0< \widehat{a}_n(\zeta_n,\delta)< 1$. We can simply set the value of $a_n$ as $\widehat{a}_n(\zeta_n,\delta)$ with $\beta_n=0$ and $\gamma_n=0$ in this case. 
    \item \textbf{Case 3:} $\widehat{a}_n(\zeta_n,\delta)\leq 0$. When $\widehat{a}_n(\zeta_n,\delta)\leq 0$, it means when $a_n=0$, $D_n(0,\zeta_n,\delta)$ is equal or better than 0. Therefore, we choose the feasible solution as $\widehat{a}_n(\zeta_n,\delta)=0$, $\gamma_n=0$, and $\beta_n=D_n(0,\zeta_n,\delta)$ which meet the condition (\ref{kktC}a).
\end{itemize}

Thus, we summarise all of these cases as follows:
\begin{align}
    \left \{\!\!
    \begin{array}{l}
         \widetilde{a}_n(\zeta_n,\delta)=\max\{\min\{\widehat{a}_n(\zeta_n,\delta),1\},0\};\\
         \widetilde{\beta}_n(\zeta_n,\delta)=\max\{D_n(0,\zeta_n,\delta),0\};\\
         \widetilde{\gamma}_n(\zeta_n,\delta)=-\min\{D_n(1,\zeta_n,\delta),0\};
    \end{array}
    \right.
\end{align}

Before we continue the analysis, one propriety of function $s_n\!\!-\!(\!1\!-\!a_n\!)S_{n,0}^{\text{max}}\!-\!a_nS_{n,1}^{\text{max}}$ which is denoted as $I_n(\zeta_n,\delta)$ for simplify in the condition (\ref{kktC}d) can be derived: \textit{$I_n(\zeta_n,\delta)$ is non-increasing of $\zeta_n$.} Then we discuss the following cased by setting $\zeta_n=0$:
\begin{itemize}
    \item \textbf{Case 1:} $I_n(0,\delta)\leq 0$. Based on the propriety mentioned above, one feasible solution meeting the condition (\ref{kktC}d) is  $\delta=0$.
    \item \textbf{Case 2:} $I_n(0,\delta)> 0$. Conversely, if $I_n(0,\delta)> 0$, the value of $\zeta_n$ can be set as $\widehat{\zeta}_n(\delta)$ based on the $I_n(\zeta_n,\delta)|_{\zeta_n=\widehat{\zeta}_n(\delta)}=0$.
\end{itemize}

The solution can be summarized as:$\widetilde{\zeta}_n(\delta)=\max\{\widehat{\zeta}_n(\delta),0\}$.

Drawing from the condition (\ref{kktS}b), the expression of $s_n$ can be deduced and we denote it as $s_n(\zeta_n,\delta)$. Similarly, we discuss the cases based on condition (\ref{kktC}c) by setting $\delta=0$:
\begin{itemize}
    \item \textbf{Case 1:}$\sum_{n\in \mathcal{N}}(\widetilde{s}_n^2(\widetilde{\zeta}_n(0),0)z_n^{(k)}+\frac{a_n^2(\widetilde{\zeta}_n(0),0)}{4z_n^{(k)}})\leq S_{e}^{\text{max}}$. In this case, we can set $\delta=0$ to meet the conditions (\ref{kktD}d) and (\ref{kktC}c).
    \item \textbf{Case 2:}$\sum_{n\in \mathcal{N}}(\widetilde{s}_n^2(\widetilde{\zeta}_n(0),0)z_n^{(k)}+\frac{a_n^2(\widetilde{\zeta}_n(0),0)}{4z_n^{(k)}})\textgreater S_{e}^{\text{max}}$. Conversely, we need to find the optimal value of the $\delta$ based on the condition (\ref{kktC}c) by using the bisection method. Then we denote the solution as $\widetilde{\delta}$.
\end{itemize}

Consequently, the optimal value of $\delta$ is expressed as:
\begin{align}
    \delta^*= \left  \{\!\!
    \begin{array}{ll}
        0, \!\!\!&  \sum_{n\in \mathcal{N}}(\widetilde{s}_n^2(\widetilde{\zeta}_n(0),0)z_n^{(k)}\!\!+\frac{a_n^2(\widetilde{\zeta}_n(0),0)}{4z_n^{(k)}})\!\leq\! S_{e}^{\text{max}},\\
        \widetilde{\delta}, \!\!\!&  \sum_{n\in \mathcal{N}}(\widetilde{s}_n^2(\widetilde{\zeta}_n(0),0)z_n^{(k)}\!\!+\frac{a_n^2(\widetilde{\zeta}_n(0),0)}{4z_n^{(k)}})\!>\! S_{e}^{\text{max}}.
    \end{array}
    \right.
\end{align}

Until now, we can get the optimal solution of variables $[\bm{a}^*,\bm{s}^*]$ and the Lagrange multipliers $[\bm{\beta}^*,\bm{\gamma}^*,\bm{\zeta}^*,\delta^*]$ by analyzing KKT conditions. $\hfill\blacksquare$

}

\end{document}